%% file: 0_Main.tex
\documentclass{svproc}
\usepackage{graphicx}
\usepackage{color}
\usepackage{amsmath,amsfonts}
\usepackage{algorithmic}
\usepackage{algorithm}
\usepackage{array,multirow}
\usepackage{textcomp}
\usepackage{siunitx}
\usepackage{url}
\usepackage{booktabs}
\usepackage{verbatim}
\usepackage{subcaption}
\usepackage[misc]{ifsym}
\usepackage{hyperref}

\renewcommand{\b}[1]{\boldsymbol{#1}} 
\newcommand{\compressParag}{\looseness=-1}
\usepackage{xcolor} 
\definecolor{myblue}{RGB}{0, 102, 204}

\begin{document}
\mainmatter              
\title{Multi-Modal Model Predictive Path Integral Control for Collision Avoidance\thanks{This research has been conducted as part of the EVENTS \& MOCO projects, which are funded by the European Union, under grant agreements No 101069614 and No 101183051. Views and opinions expressed are, however, those of the authors only and do not necessarily reflect those of the European Union or European Commission. Neither the European Union nor the granting authority can be held responsible for them.\compressParag}}
\titlerunning{Multi-Modal MPPI for Collision Avoidance}  
%
\author{Alberto Bertipaglia\inst{1}\textsuperscript{(\Letter)}\textsuperscript{[0000-0003-0364-8833]} \and Dariu M. Gavrila\inst{1}\textsuperscript{[0000-0002-1810-4196]} \and
Barys Shyrokau\inst{1}\textsuperscript{[0000-0003-4530-8853]}}
\authorrunning{Alberto Bertipaglia et al.} 
%
%
\institute{Delft University of Technology, Delft 2628 CD, The Netherlands\\
\email{\{a.bertipaglia, d.m.gavrila, b.shyrokau\}@tudelft.nl}}

\maketitle              

\begin{abstract}
This paper proposes a novel approach to motion planning and decision-making for automated vehicles, using a multi-modal Model Predictive Path Integral control algorithm. The method samples with Sobol sequences around the prior input and incorporates analytical solutions for collision avoidance. By leveraging multiple modes, the multi-modal control algorithm explores diverse trajectories, such as manoeuvring around obstacles or stopping safely before them, mitigating the risk of sub-optimal solutions. A non-linear single-track vehicle model with a Fiala tyre serves as the prediction model, and tyre force constraints within the friction circle are enforced to ensure vehicle stability during evasive manoeuvres. The optimised steering angle and longitudinal acceleration are computed to generate a collision-free trajectory and to control the vehicle. In a high-fidelity simulation environment, we demonstrate that the proposed algorithm can successfully avoid obstacles, keeping the vehicle stable while driving a double lane change manoeuvre on high and low-friction road surfaces and occlusion scenarios with moving obstacles, outperforming a standard Model Predictive Path Integral approach.\compressParag 
\end{abstract}
%
%
\input{1_Introduction}
\input{2_MPPI}
\input{3_Experiments}
\input{4_Results}
\input{5_Conclusions}
\bibliographystyle{splncs04}
\bibliography{references.bib}
\end{document}

%% file: 1_Introduction.tex
\section{Introduction}
\label{introduction}
The ability to design and select a safe trajectory during an evasive manoeuvre at high and low-friction conditions is a critical factor for the success of automated vehicles. Tyre nonlinearities, combined with the complexity of dynamic environments and the uncertainty of the road friction coefficient, make this problem particularly challenging \cite{bertipaglia2023model,de2024topology}. The complexity of motion planning also arises from the two interconnected tasks which must be addressed. First, it involves making a high-level decision on how to avoid a potential collision, such as overtaking the obstacle (from the right or left side) or stopping the vehicle before reaching the obstacle. Second, it requires performing local (low-level) motion planning, generating a trajectory that avoids collisions with obstacles while adhering to vehicle stability constraints and actuator feasibility \cite{testouri2023towards}.\newline\indent
A typical approach involves explicitly addressing these two levels of motion planning by developing a Topology-Driven Model Predictive Control (T-MPC) framework \cite{de2024topology}. The T-MPC is divided into two components: a global or guidance planner based on Visibility-Probabilistic Road Maps \cite{simeon2000visibility}, which computes simple trajectories within different homotopy classes, and a local planner based on Model Predictive Control (MPC), which optimises each trajectory in parallel. While the T-MPC successfully mitigates the issue of local optimality often encountered in optimisation algorithms, it has certain limitations. The vehicle dynamics is oversimplified, and the need to run a local motion planner to optimise multiple trajectories in real-time restricts the complexity of the prediction model to ensure real-time feasibility. These limitations are particularly problematic at low-friction conditions, where inaccuracies in the prediction model significantly impact trajectory optimisation \cite{bertipaglia2023model}. To address the challenges of solving high-dimensional, non-linear optimisation problems in real-time, a technique called Model Predictive Path Integral Control (MPPI) \cite{testouri2023towards} has recently demonstrated its effectiveness in motion planning for automated vehicles. MPPI employs a sampling-based strategy to determine the safest trajectory by forward sampling and simulating many trajectories, which are then evaluated based on a cost function. This evaluation can be performed efficiently using parallel processing. The MPPI approach inherently explores a wide range of trajectories, including those valid also with model inaccuracies and road friction variation. Standard MPPI samples control inputs from a Gaussian distribution, using the previously computed input sequence as the mean. However, this approach may lead the algorithm to converge to local minima, limiting its ability to explore alternative control input sequences. One solution is to bias the MPPI sampling using ancillary controllers \cite{trevisan2024biased}. Another extends MPPI by propagating both mean and covariance of the system dynamics via the unscented transform \cite{mohamed2025towards}, and can be extended with probabilistic collision checking \cite{mohamed2025chance}. While this enhances state-space exploration, its performance deteriorates as system dimensionality increases.\newline\compressParag
This paper proposes a novel multi-modal MPPI approach that samples trajectories using a Sobol sequence and incorporates both prior inputs and analytical collision avoidance solutions as means. This enables the MPPI to explore diverse trajectories, preventing the algorithm from achieving only local minimum solutions. Additionally, the objective function of each mode is customised to align with its corresponding analytical solution. For example, the vehicle can brake and stop before colliding with an obstacle without incurring a significant velocity tracking error. The prediction model is a nonlinear single-track vehicle model with a Fiala tyre formulation, and tyre force constraints are enforced via the friction circle to ensure stability. We assess the proposed approach's performance in a high-fidelity simulation environment, optimising the trajectory of a double lane change in high and low-friction conditions and an occlusion scenario with moving obstacles.\compressParag\newline\indent
The contributions of this paper are twofold. First, it introduces the first multi-modal MPPI, which samples trajectories using a Sobol sequence and incorporates both prior input sequences and analytical solutions to collision avoidance problems as means. This enables the proposed approach to successfully perform collision avoidance manoeuvres without colliding with obstacles, outperforming standard MPPI methods for motion planning \cite{testouri2023towards}. Second, it presents a motion planner capable of executing evasive manoeuvres up to the vehicle’s handling limits, using a nonlinear single-track model with Fiala tyre dynamics and friction-circle constraints. In contrast, standard methods remain limited to the linear operating regime \cite{de2025vehicle,testouri2023towards}.

%% file: 2_MPPI.tex
\section{Multi-Modal Model Predictive Path Integral Control}
\label{Multi_Modal_MPPI}
This section explains how the Multi-Modal MPPI is formulated and proposed.
\subsection{Model Predictive Path Integral Control}
\label{MPPI}
MPPI control is a sampling-based stochastic optimal control algorithm to solve a non-linear optimisation problem subject to non-linear dynamics and non-convex constraints \cite{williams2016aggressive,williams2017model}. The MPPI utilises Monte Carlo sampling to explore a large number of control sequences, which are propagated through the discrete-time model $\left(f\right)$ over a finite time horizon, and the obtained state trajectories' performance is evaluated using a cost function $\left(J\right)$. The computed cost for state trajectories is used to weight each sampled control sequence (or rollout), forming a path integral estimate of the optimal control input. The weight of each rollout is computed as follows:
\begin{equation}
    \omega_k = \frac{1}{\eta}\text{exp}\left(-\frac{1}{\lambda}\left(S_k-\rho\right)\right)\text{, where }\eta=\sum_{k=1}^{K}{\text{exp}\left(-\frac{1}{\lambda}\left(S_k-\rho\right)\right)}
    \label{eq:weights}
\end{equation}
where $\omega$ is the weight for each rollout $k\in[1,K]$, $K$ is the total number of rollouts, $\eta$ is the normalization constant to ensure $\sum_{k=1}^{K}{\omega_k}=1$, $S_k$ is the trajectory cost for each rollout, and $\rho$ is the minimum cost $S_k$ which is used to avoids large exponents that could cause underflow. Regarding $\lambda$, it is the temperature parameter which regulates the selectivity of the weighting, a small $\lambda$ prioritises the lowest-cost trajectories, so it favours exploitation, vice versa, a large $\lambda$ favours exploration, flattening the cost rollouts \cite{zhang2024multi}. Finally, the sampling weights are used to approximate the optimal control sequence, which is propagated to the discrete-time model $f$ to compute the optimum collision-free trajectory. \newline\noindent
A core element of sampling-based algorithms is how the $K$ control rollouts are sampled from a stochastic distribution because it is the only tool that the MPPI have to explore different solutions and trajectories \cite{park2025csc}. MPPI relies critically on the efficiency and quality of trajectory exploration. Thus, we propose to use Sobol sequences, a class of quasi-random low-discrepancy sequences, which offer several advantages over conventional stochastic sampling techniques. First, Sobol sequences provide more uniform coverage of the sampling space, especially in high-dimensional settings, than pseudo-random Gaussian sampling \cite{testouri2023towards,OutputMPPI}. Second, Sobol sequences exhibit superior scalability with dimensionality compared to other low-discrepancy sequences such as Halton sequences \cite{williams2017model,zhang2024multi}, making them well-suited for motion planning in complex multi-input systems with long prediction horizons. The problem we aim to solve is formally expressed as follows:
\begin{subequations}
    \label{eq:mppi}
    \begin{align}
        \min_{\b{u} \in \mathbb{U}, \b{x}\in\mathbb{X}} \hspace{1em} & \sum_{t = 1}^T J(\b{x}_t, \b{u}_t)\\
            \textrm{s.t.} \hspace{2em} & \b{x}_0 = \b{x}_{\textrm{init}} \label{eq:init_condition} \\
            & \b{x}_{t + 1} = f(\b{x}_t, \b{u}_t), \ t = 0, \hdots, T-1 \label{eq:dynamics}
    \end{align}
\end{subequations}
where $T$ is the length of the prediction horizon (50 steps), $\b{x}$ and $\b{u}$ are, respectively, the vehicle states and the vehicle inputs. The objective function $J$ includes path tracking and motion planning costs \cite{bertipaglia2023model}, velocity error term, and penalties on the inputs \cite{bertipaglia2023model}. The initial conditions are imposed by eq. \ref{eq:init_condition}, and the vehicle dynamics by eq. \ref{eq:dynamics}.
\subsection{Vehicle Prediction Model}
\label{VehicleModel}
A non-linear single-track vehicle model is adopted in the Multi-Modal MPPI. The vehicle state vector $\left(x = [X, Y, \phi, v_x, v_y, r, \theta, \delta, a_x]\right)$ describes the position and orientation of the vehicle's centre of gravity (CoG) in a Cartesian frame, including longitudinal $\left(X \right)$, lateral $\left(Y \right)$ positions, and heading angle $\left(\psi \right)$. Velocity states consist of longitudinal $\left(v_x\right)$ and lateral $\left(v_y\right)$ velocities, yaw rate $\left(\dot{r}\right)$, and the travelled distance $\left(\theta\right)$, the latter is used to evaluate position relative to a reference path \cite{bertipaglia2023model,BertipagliaTVJ}. Steering angle $\left(\delta\right)$ and longitudinal acceleration $\left(\dot{a_x}\right)$ are integrated from the control inputs. The vehicle dynamics is represented as follows:\compressParag
\begin{equation}
    \begin{split}
    \begin{cases}
        \dot{X} = v_x \cos \left(\psi\right) - v_y \sin \left(\psi\right)\\
        \dot{Y} = v_x \sin \left(\psi\right) + v_y \cos \left(\psi\right)\\
        \dot{\psi} = r\\
        \dot{v}_x = \frac{- F_{yf}\left(x, u_v\right) \sin \left(\delta \right) + F_{xf}\left(x, u_v\right) \cos \left(\delta\right) + F_{xr}\left(x, u_v\right) - F_{drag}}{m} + r v_y\\
        \dot{v}_y = \frac{F_{yf}\left(x, u_v\right) \cos \left(\delta \right) + F_{xf} \sin \left(\delta\right) + F_{yr}\left(x, u_v\right)}{m} - r v_x\\
        \dot{r} = \frac{l_f F_{yf}\left(x, u_v\right) \cos \left(\delta \right) +  l_f F_{xf}\left(x, u_v\right) \sin \left(\delta\right) - l_r F_{yr}\left(x, u_v\right) }{I_{zz}}\\
        \dot{\theta} = \sqrt{v_x^2 + v_y^2}
    \end{cases}
    \end{split}
    \label{eq:Single}
\end{equation}
where $F_{xi}$ and $F_{yi}$  are, respectively, the longitudinal and lateral tyre forces, $i$ stands for front $\left(f\right)$ or rear $\left(r\right)$, $l_f$ and $l_r$ are the distance from the CoG to axles, $I_{zz}$ is the vehicle yaw inertia, $m$ is the vehicle mass and $F_{drag}$ is the aerodynamic drag resistance. The vehicle model inputs $\left(u_v\right)$ are the road-wheel angle rate $(\dot{\delta})$, and the longitudinal jerk applied at the CoG $(\dot{a}_x)$. The control input rates are integrated into the prediction model before being applied to the vehicle. A Fiala tyre model computes the lateral tyre force for each axle, while the longitudinal force is computed using an input of the system \cite{bertipaglia2023model}. The non-linear coupling between the tyre longitudinal and lateral tyre forces is captured via the friction circle \cite{BertipagliaTVJ}. The tyre parameters are optimised through quasi-steady-state circular driving in a high-fidelity simulation based on a Delft-Tyre model 6.1. Vehicle and obstacles are represented by circles for efficient computation of Euclidean distances: two circles model the vehicle, and a single circle represents each obstacle.\compressParag
\subsection{Cost Function}
\label{CostFunction}
The proposed Multi-Modal MPPI utilises a non-linear cost function to iteratively solve an optimal control problem, enabling the vehicle to drive at the limit of handling while avoiding obstacles. The cost function objectives include maintaining safe distances from obstacles and road edges, tracking a reference mission path, optimising longitudinal velocity, ensuring stability by limiting yaw rate and sideslip angle \cite{bertipagliaUKF}, and enforcing physical actuator constraints (maximum steering angle and acceleration). The proposed cost function $\left(J\right)$ is:
\begin{equation}
    \begin{split}
        J = \sum_{i=1}^{T} \biggl( & q_{e_{Con}} e_{Con, i}^2 + q_{e_{Lag}} e_{Lag, i}^2 + q_{e_{Vel}} e_{Vel} + q_{\dot{\delta}} \dot{\delta_i}^2 + q_{\dot{a_x}} \dot{a_{x_i}}^2 +\\
        & + q_{\delta} \left(|\delta_i|>\delta_{max}\right) + q_{a_x} \left(|a_{x_i}|>a_{x_{max}}\right) + q_{\beta} \left(|\beta_i|>\beta_{max}\right) + \\
        & + q_{r} \left(|r_i|>r_{max}\right) + q_{Tf} \left(F_{xj_i}> s_c\mu_jF_{zj_i}\right) + q_{St} \left(v_x< v_{x_{min}}\right) + \\
        & +\sum_{j=1}^{N_{obs}}\left( q_{e_{V2O}} e_{V2O, j, i}^2 \right) + \sum_{j=1}^{N_{edg}}\left( q_{e_{V2E}} e_{V2E, j, i}^2 \right) \biggr) 
    \end{split}
    \label{eq:cost}
\end{equation}
where $T$ is the prediction horizon, $N_{obs}$ the number of obstacles, $N_{edg}=2$ the road edges, and $q_*$ the tuning weights. These are optimised to minimise longitudinal velocity errors and reduce sideslip peaks \cite{BertipagliaTVJ}. The reference mission path is tracked using contouring $\left(e_{Con}\right)$ and lag errors $\left(e_{Lag}\right)$ \cite{bertipaglia2023model}. Velocity tracking uses a Log-Cosh penalty, defined as:\compressParag
\begin{equation}
    \label{eq:velocity}
    e_{Vel} = \log \left( \cosh{ \left(v_x - v_{x, des} \right)} \right)
\end{equation}
where $v_x$ is the vehicle’s longitudinal velocity and $v_{x,des}$ the desired velocity from the mission planner. Eq. \ref{eq:velocity} blends quadratic and linear penalties, enabling smooth tracking while tolerating larger deviations for collision avoidance. Input smoothness and feasibility are enforced via penalties on input rates $\left(\dot\delta,, \dot a_x\right)$. Actuator limits are treated as soft constraints due to MPPI limitations, with maximum steering and acceleration set by $\left(\delta_{max},, \dot a_{x_{max}}\right)$. The cost terms related to yaw rate $r$, sideslip angle $\beta$, and longitudinal tyre forces $F_{xj}$ ensure vehicle stability by limiting the maximum allowable yaw rate $r_{max}$ and sideslip angle $\beta_{max}$. Stability is maintained by restricting the total tyre force available at each axle according to the tyre friction circle, represented by the weight $q_{Tf}$. Specifically, longitudinal force $F_{xj}$ is constrained by $F_{xj}<S_c\mu F_{zj}$, where $\mu$ is the road friction coefficient. Due to estimation uncertainties, a safety factor $S_c$ of 0.95 is applied to reduce the maximum allowable longitudinal force.\newline\noindent
The controller dynamically adjusts the mission planner trajectory to maintain safe distances from obstacles and road edges. It calculates the vehicle-to-obstacle (V2O) distance error as $\left(e_{V20} = D_{V2O} - D_{Sft, O}\right)$, where $D_{V2O}$ is the actual distance and $D_{Sft, O}$ is a predefined safe distance. If $D_{V2O}$ exceeds $D_{Sft, O}$, the associated weight $q_{V2O}$ is zero. However, when $D_{V2O}$ falls below the safety threshold, the error is penalised to facilitate a safe margin. A similar logic applies to maintaining safe distances from road edges, using the vehicle-to-edge error $\left(e_{V2E} = D_{V2E} - D_{Sft, E}\right)$, where $D_{V2E}$ and $D_{Sft, E}$ represent the actual and safe distances to the road edge, respectively.
\subsection{Multi-Modal Sampling}
\label{Inputs}
A standard MPPI algorithm \cite{testouri2023towards} utilises a single stochastic distribution centred around the previously optimised control sequence, shifted forward in time. Despite its effectiveness in many applications, this approach often concentrates sampled trajectories within high-cost regions or lacks sufficient diversity, making MPPI susceptible to local minima and increasing the risk of collisions with obstacles \cite{park2025csc}. This occurs because standard MPPI tends to become trapped in local minima, struggling to alter its decisions quickly enough to avoid obstacles.\compressParag\newline\noindent
To address this issue, we propose a Multi-Modal MPPI algorithm that samples around four distinct solutions. The first follows the previously optimised control input sequence, as in standard MPPI. The other three are activated when the Time to Closest Point of Approach (TCPA) falls below \SI{2}{s}, and correspond to maximum braking, maximum acceleration, and an evasive manoeuvre based on the wary approach \cite{fors2020autonomous}. By concurrently exploring multiple strategies, our method increases trajectory diversity and reduces sensitivity to local minima. Each mode computes its own set of importance weights, which are combined to produce a final control sequence, allowing smooth transitions between strategies \cite{zhang2024multi}.

%% file: 3_Experiments.tex
\section{Experimental Setup}
\label{experiments}
The proposed approach is implemented on a PC equipped with an Intel Xeon W-2223 quad-core CPU, 32 GB RAM, and an NVIDIA TITAN X (Pascal) GPU. The Multi-Modal MPPI algorithm runs on the GPU, while the high-fidelity vehicle model works on the CPU. The prediction model is discretised using a Runge-Kutta 2 method \cite{bertipaglia2023model}. A sampling time of \SI{0.05}{s}, a prediction horizon of \SI{50}{} steps, and a total amount of 2600 samples are chosen to ensure real-time feasibility. The optimisation problem involves parallel evaluation of numerous trajectories generated by Monte Carlo simulations. To exploit this inherent parallelism, the Multi-Modal MPPI controller is developed in MATLAB and deployed onto the GPU using CUDA via MATLAB GPU Coder. This setup achieves an average solving time of \SI{25.4}{ms} and a maximum of \SI{31.6}{ms}, although mathematical guarantees for convergence within real-time constraints are not provided. The vehicle plant model runs independently at \SI{1000}{Hz} on a separate CPU core. It employs a high-fidelity third-generation Toyota Prius vehicle model integrated with the IPG CarMaker simulation platform. Tyre dynamics are modelled using Delft-Tyre 6.1, and actuator dynamics are represented via second-order transfer functions to enhance simulation accuracy \cite{bertipaglia2023model}.\compressParag

%% file: 4_Results.tex
\section{Results}
\label{results}
This section presents the performance of the proposed Multi-Modal MPPI approach compared to a standard MPPI baseline \cite{testouri2023towards} in various driving scenarios.
\subsection{Double Lane Change with High-Friction Conditions}
\begin{figure*}[!t]
    \centering
    \begin{subfigure}{0.32\columnwidth}
        \centering
        \includegraphics[width=1\columnwidth]{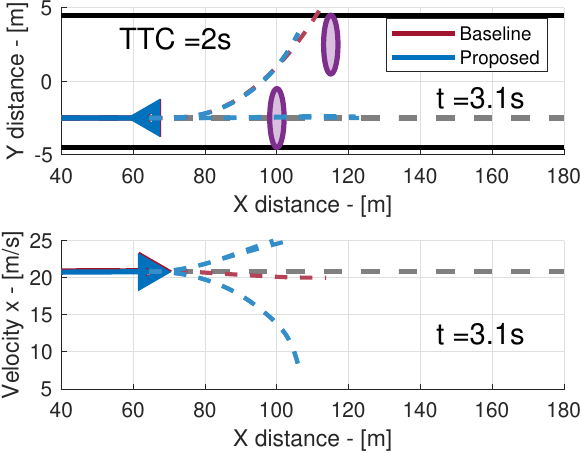}
        \caption{\centering}
        \label{fig:DL_57_init}
    \end{subfigure} \hfill
    \begin{subfigure}{0.32\columnwidth}
        \centering
        \includegraphics[width=1\columnwidth]{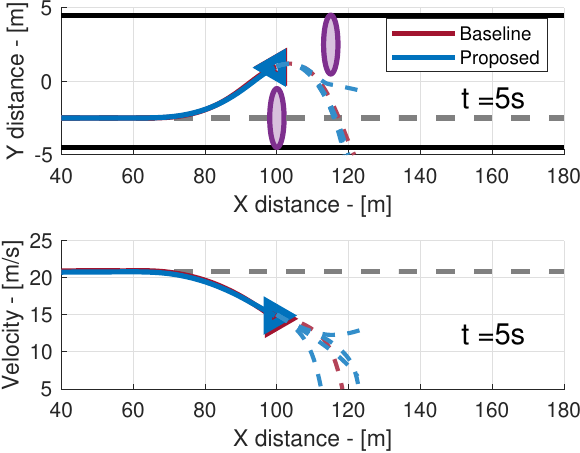}`
        \caption{\centering}
        \label{fig:DL_57_medium}
    \end{subfigure} \hfill
    \begin{subfigure}{0.32\columnwidth}
        \centering
        \includegraphics[width=1\columnwidth]{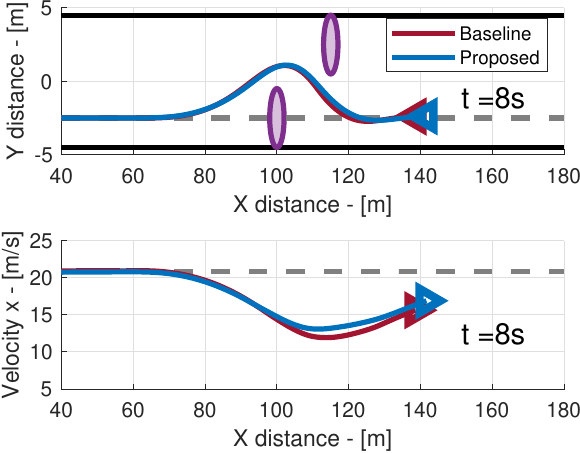}
        \caption{\centering}
        \label{fig:DL_57_end}
    \end{subfigure}\\
    \caption{Double lane change under high-friction conditions with TTC $= \SI{2}{s}$.}  
    \label{fig:DL_57}
\end{figure*}

\begin{figure*}[!t]
    \centering
    \begin{subfigure}{0.32\columnwidth}
        \centering
        \includegraphics[width=1\columnwidth]{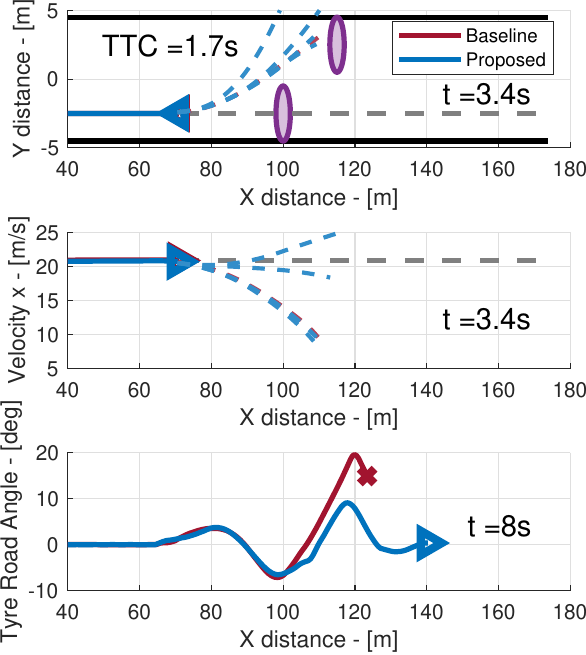}
        \caption{\centering}
        \label{fig:H_64_init}
    \end{subfigure} \hfill
    \begin{subfigure}{0.32\columnwidth}
        \centering
        \includegraphics[width=1\columnwidth]{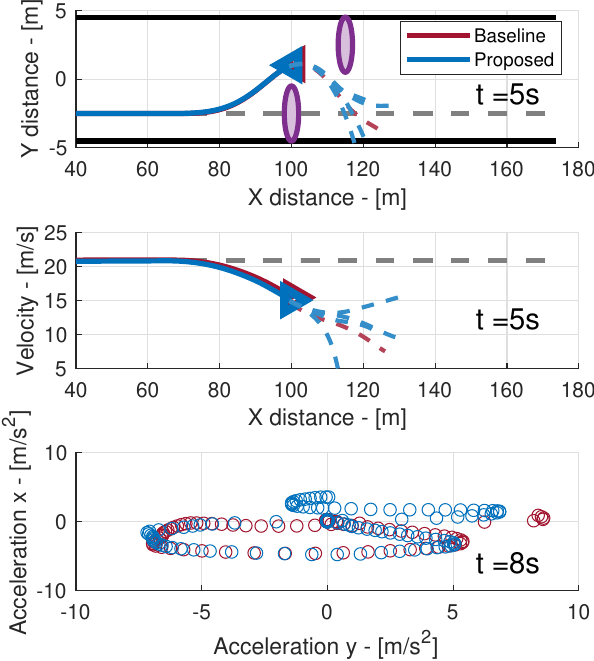}
        \caption{\centering}
        \label{fig:H_64_medium}
    \end{subfigure} \hfill
    \begin{subfigure}{0.32\columnwidth}
        \centering
        \includegraphics[width=1\columnwidth]{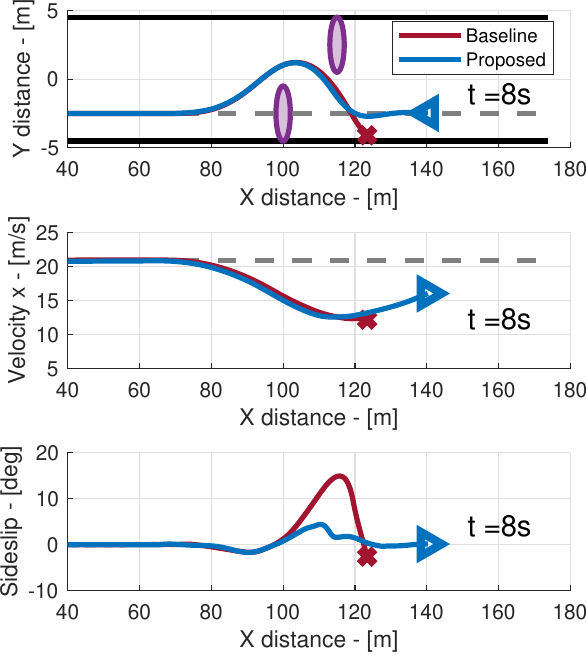}
        \caption{\centering}
        \label{fig:H_64_end}
    \end{subfigure}
    \caption{Double lane change under high-friction conditions with TTC $= \SI{1.7}{s}$.} 
    \label{fig:DL_64}
\end{figure*}

Fig.\ref{fig:DL_57} shows the planned and executed vehicle trajectories for both the proposed Multi-Modal MPPI and the standard MPPI baseline \cite{testouri2023towards}. In this scenario, the vehicle encounters two static obstacles that appear with a time-to-collision (TTC) of \SI{2}{s}, requiring a double lane change. Both planners successfully avoid collision, but they achieve this through different decision-making processes. As shown in Fig.\ref{fig:DL_57_init}, when the obstacles first appear, the baseline planner immediately plans an evasive manoeuvre. In contrast, the proposed Multi-Modal MPPI evaluates multiple strategies, including both an evasive manoeuvre and a harsh braking to stop the vehicle before the first obstacle. Figs.\ref{fig:DL_57_medium} and \ref{fig:DL_57_end} demonstrate that after evaluating the feasibility of braking, the proposed planner selects a similar evasive manoeuvre to the baseline, successfully completing the double lane change.\compressParag\newline\noindent
Fig.\ref{fig:DL_64} shows the same double lane change scenario, now with obstacles appearing at a reduced TTC of \SI{1.7}{s}. In this case, the proposed planner excludes a full stop due to insufficient braking distance but considers both harsh braking and an evasive manoeuvre, enabling broader cost function exploration. As a result, the vehicle decelerates more before initiating the lane change. In contrast, the baseline planner gets trapped in a local minimum, failing to decelerate adequately before steering. As seen in Fig.\ref{fig:H_64_medium}, it initiates the second lane change at a higher speed without further braking, compromising stability. Consequently, the baseline exits the road boundaries (Fig.\ref{fig:H_64_end}). The proposed Multi-Modal MPPI, by dynamically adapting its strategy, completes the manoeuvre successfully and demonstrates superior performance.\compressParag
\subsection{Obstacle Occlusion Scenario}
\begin{figure*}[!t]
    \centering
    \begin{subfigure}{0.32\columnwidth}
        \centering
        \includegraphics[width=1\columnwidth]{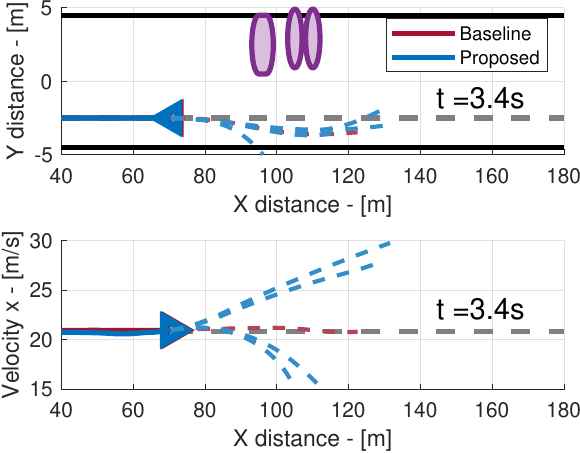}
        \caption{\centering}
        \label{fig:H_Acc_init}
    \end{subfigure} \hfill
    \begin{subfigure}{0.32\columnwidth}
        \centering
        \includegraphics[width=1\columnwidth]{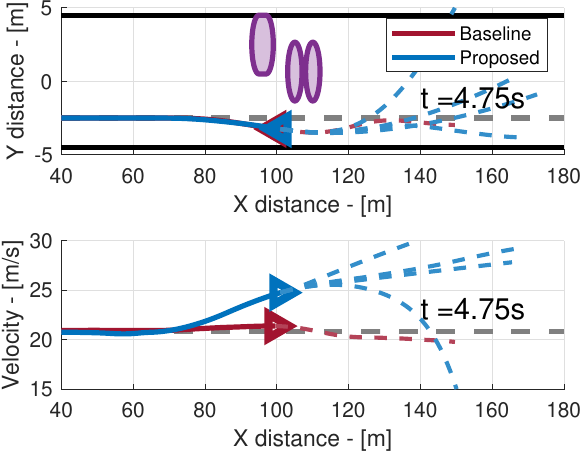}
        \caption{\centering}
        \label{fig:H_Acc_medium}
    \end{subfigure} \hfill
    \begin{subfigure}{0.32\columnwidth}
        \centering
        \includegraphics[width=1\columnwidth]{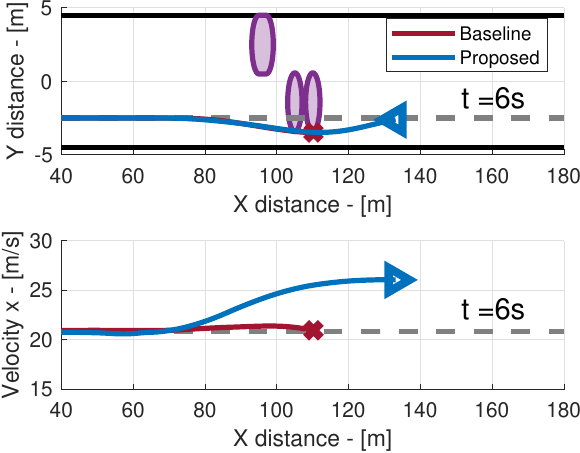}
        \caption{\centering}
        \label{fig:H_Acc_end}
    \end{subfigure}
    \caption{Obstacle occlusion scenario under high-friction conditions.}
    \label{fig:H_Acc}
\end{figure*} 
Fig.\ref{fig:H_Acc} shows the vehicle trajectories in an occlusion scenario. A static barrier occludes the perception of two crossing dynamic obstacles, simulating vulnerable road users. This scenario presents a critical situation where the vehicle cannot come to a full stop or execute a typical evasive manoeuvre. Instead, acceleration becomes the only viable strategy to avoid a collision. Initially, as shown in Fig.\ref{fig:H_Acc_init}, both planners attempt to decelerate in response to the perceived obstacles. However, as it becomes clear that stopping is not feasible, the proposed Multi-Modal MPPI avoids being trapped in a local minimum. Fig.\ref{fig:H_Acc_medium} shows that the proposed planner shifts strategy and begins accelerating to safely pass in front of the crossing obstacles. In contrast, the baseline planner fails to adapt its plan in time, remaining in a suboptimal deceleration mode. Fig.\ref{fig:H_Acc_end} confirms that only the proposed approach successfully completes the manoeuvre, while the baseline fails to avoid the collision. This result highlights the advantage of multi-modal sampling, where alternative strategies must be considered to ensure safety.\compressParag
\subsection{Double Lane Change with Low-Friction Conditions}
\begin{figure*}[!t]
    \centering
    \begin{subfigure}{0.32\columnwidth}
        \centering
        \includegraphics[width=1\columnwidth]{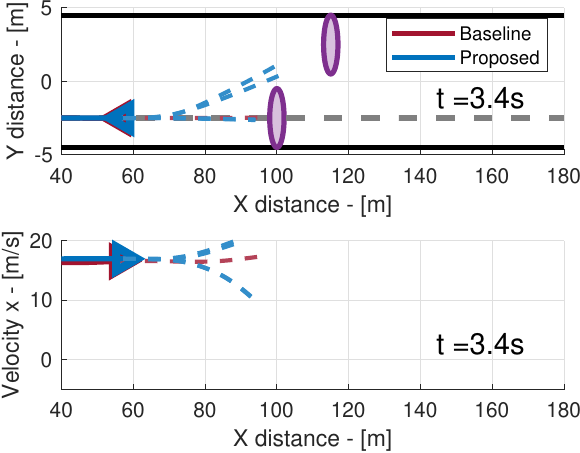}
        \caption{\centering}
        \label{fig:L_57_init}
    \end{subfigure} \hfill
    \begin{subfigure}{0.32\columnwidth}
        \centering
        \includegraphics[width=1\columnwidth]{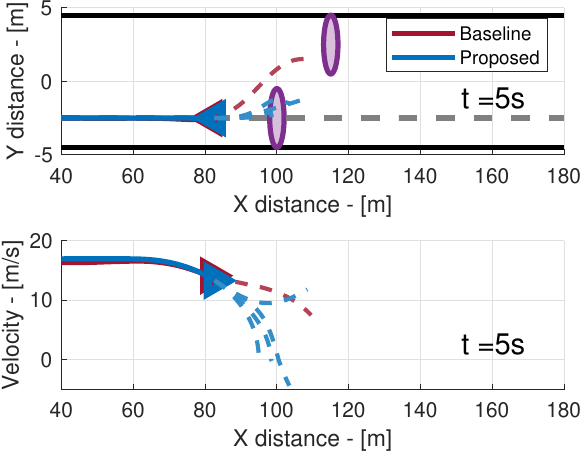}
        \caption{\centering}
        \label{fig:L_57_medium}
    \end{subfigure} \hfill
    \begin{subfigure}{0.32\columnwidth}
        \centering
        \includegraphics[width=1\columnwidth]{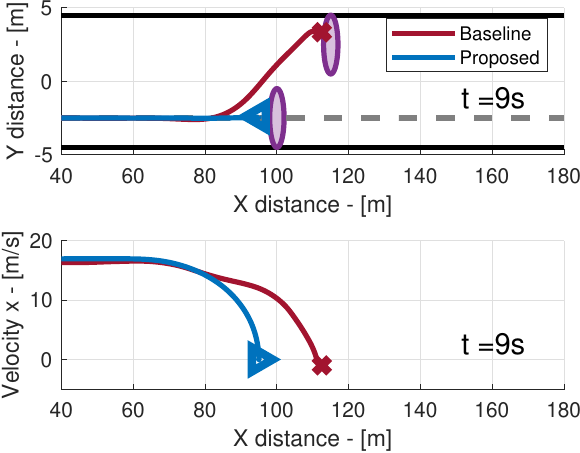}
        \caption{\centering}
        \label{fig:L_57_end}
    \end{subfigure}
    \caption{Double lane change under low-friction conditions.}    
    \label{fig:L_57}
\end{figure*} 
Fig.\ref{fig:L_57} presents the same double lane change scenario, but under low-friction conditions, and a vehicle's initial velocity of \SI{60}{km/h}. The proposed planner is the only one to consider a harsh braking action in front of the first obstacle. Fig.\ref{fig:L_57_medium} reveals a key difference in behaviour. The proposed Multi-Modal MPPI decides to brake aggressively and brings the vehicle to a complete stop before reaching the first obstacle. Vice versa, the baseline planner initiates an evasive manoeuvre without sufficient deceleration. Due to the reduced friction, the baseline loses stability and collides with the second obstacle during the lane change. Fig.\ref{fig:L_57_end} confirms that the proposed planner successfully avoids both obstacles by stopping the vehicle in time. This demonstrates the benefit of including a harsh braking strategy among the sampled control modes, especially in low-adhesion scenarios.\compressParag

%% file: 5_Conclusions.tex
\section{Conclusions}
\label{Conclusion}
This paper presented a novel approach to motion planning and decision-making for automated vehicles using a Multi-Modal MPPI. By sampling input trajectories via Sobol sequences centred around both previously optimised inputs and analytical solutions, the method explores multiple minima of the cost function and adapts to complex scenarios. In a double lane change manoeuvre under high-friction conditions with a TTC of \SI{2}{s}, both the proposed and baseline planners generate similar evasive trajectories. However, the proposed method evaluates a broader set of alternatives, including harsh braking. This proves critical when TTC is reduced to \SI{1.7}{s}, where only the proposed planner decelerates sufficiently to complete the manoeuvre safely, unlike the baseline, which fails to stay in the road boundaries. Similar advantages are seen in scenarios with occluded obstacles and under low-friction conditions, where the proposed method adapts via acceleration and braking, while the baseline remains stuck in local minima. Future work will involve validation on a real vehicle at a test track, with results to be presented at the conference.\compressParag

%% file: 0_Main.bbl
\begin{thebibliography}{10}
\providecommand{\url}[1]{\texttt{#1}}
\providecommand{\urlprefix}{URL }
\providecommand{\doi}[1]{https://doi.org/#1}

\bibitem{bertipaglia2023model}
Bertipaglia, A., Alirezaei, M., Happee, R., Shyrokau, B.: Model predictive contouring control for vehicle obstacle avoidance at the limit of handling. In: Symposium on the Dynamics of Vehicle on Roads and on Tracks (2023)

\bibitem{bertipagliaUKF}
Bertipaglia, A., Alirezaei, M., Happee, R., Shyrokau, B.: An unscented kalman filter-informed neural network for vehicle sideslip angle estimation. IEEE Transactions on Vehicular Technology  \textbf{73}(9),  12731--12746 (2024)

\bibitem{BertipagliaTVJ}
Bertipaglia, A., Tavernini, D., Montanaro, U., Alirezaei, M., Happee, R., Sorniotti, A., Shyrokau, B.: On the benefits of torque vectoring for automated collision avoidance at the limits of handling. IEEE Transactions on Vehicular Technology  \textbf{74}(6),  8756--8771 (2025)

\bibitem{fors2020autonomous}
Fors, V., Olofsson, B., Nielsen, L.: Autonomous wary collision avoidance. IEEE Transactions on Intelligent Vehicles  \textbf{6}(2),  353--365 (2020)

\bibitem{de2025vehicle}
de~Groot, O., Bertipaglia, A., Boekema, H., Jain, V., Kegl, M., Kotian, V., Lentsch, T., Lin, Y., Messiou, C., Schippers, E., et~al.: A vehicle system for navigating among vulnerable road users including remote operation. In: IEEE Intelligent Vehicles Symposium (2025)

\bibitem{de2024topology}
de~Groot, O., Ferranti, L., Gavrila, D.M., Alonso-Mora, J.: Topology-driven parallel trajectory optimization in dynamic environments. IEEE Transactions on Robotics  \textbf{41},  110--126 (2025)

\bibitem{mohamed2025chance}
Mohamed, I.S., Ali, M., Liu, L.: Chance-constrained sampling-based mpc for collision avoidance in uncertain dynamic environments. IEEE Robotics and Automation Letters  \textbf{10}(7),  7492--7499 (2025)

\bibitem{mohamed2025towards}
Mohamed, I.S., Xu, J., Sukhatme, G.S., Liu, L.: Towards efficient mppi trajectory generation with unscented guidance: U-mppi control strategy. IEEE Transactions on Robotics  \textbf{41},  1172--1192 (2025)

\bibitem{park2025csc}
Park, L., Jang, K., Kim, S.: Csc-mppi: A novel constrained mppi framework with dbscan for reliable obstacle avoidance. arXiv preprint arXiv:2506.16386  (2025)

\bibitem{simeon2000visibility}
Sim{\'e}on, T., Laumond, J.P., Nissoux, C.: Visibility-based probabilistic roadmaps for motion planning. Advanced Robotics  \textbf{14}(6),  477--493 (2000)

\bibitem{testouri2023towards}
Testouri, M., Elghazaly, G., Frank, R.: Towards a safe real-time motion planning framework for autonomous driving systems: A model predictive path integral approach. In: IEEE International Conference on Robotics, Automation and Artificial Intelligence. pp. 231--238 (2023)

\bibitem{trevisan2024biased}
Trevisan, E., Alonso-Mora, J.: Biased-mppi: Informing sampling-based model predictive control by fusing ancillary controllers. IEEE Robotics and Automation Letters  \textbf{9}(6),  5871--5878 (2024)

\bibitem{williams2017model}
Williams, G., Aldrich, A., Theodorou, E.A.: Model predictive path integral control: From theory to parallel computation. Journal of Guidance, Control, and Dynamics  \textbf{40}(2),  344--357 (2017)

\bibitem{williams2016aggressive}
Williams, G., Drews, P., Goldfain, B., Rehg, J.M., Theodorou, E.A.: Aggressive driving with model predictive path integral control. In: IEEE International Conference on Robotics and Automation. pp. 1433--1440 (2016)

\bibitem{OutputMPPI}
Yan, L.L., Devasia, S.: Output-sampled model predictive path integral control (o-mppi) for increased efficiency. In: IEEE International Conference on Robotics and Automation. pp. 14279--14285 (2024)

\bibitem{zhang2024multi}
Zhang, Y., Pezzato, C., Trevisan, E., Salmi, C., Corbato, C.H., Alonso-Mora, J.: Multi-modal mppi and active inference for reactive task and motion planning. IEEE Robotics and Automation Letters  \textbf{9}(9),  7461--7468 (2024)

\end{thebibliography}
